\documentclass[10pt,twocolumn,letterpaper]{article}

\usepackage{cvpr}
\usepackage{times}
\usepackage{epsfig}
\usepackage{graphicx,caption}
\usepackage{amsmath}
\usepackage{amssymb}
\usepackage[shortlabels]{enumitem}
\usepackage{threeparttable}

\usepackage{url}
\usepackage[english]{babel}
\usepackage{amsfonts}
\usepackage{subfigure}

\newcommand{\vpara}[1]{\vspace{0.1in}\noindent\textbf{#1}}

\usepackage[pagebackref=true,breaklinks=true,letterpaper=true,colorlinks,bookmarks=false]{hyperref}

\cvprfinalcopy 

\def\cvprPaperID{2100} 

 \ifcvprfinal\fi
\begin{document}

\title{Stacked Generative Adversarial Networks}


\author{Xun Huang$^{1}$ \qquad Yixuan Li$^{2}$ \qquad Omid Poursaeed$^{2}$  \qquad John Hopcroft$^{1}$ \qquad Serge Belongie$^{1,3}$ \\
	$^1${Department of Computer Science, Cornell University}\\
    $^2${School of Electrical and Computer Engineering, Cornell University}\quad
	$^3${Cornell Tech}\\
	{\tt\small \{xh258,yl2363,op63,sjb344\}@cornell.edu jeh@cs.cornell.edu}
}

\maketitle

\begin{abstract}
In this paper, we propose a novel generative model named Stacked Generative Adversarial Networks (SGAN), which is trained to invert the hierarchical representations of a bottom-up discriminative network. Our model consists of a top-down stack of GANs, each learned to generate lower-level representations conditioned on higher-level representations. A representation discriminator is introduced at each feature hierarchy to encourage the representation manifold of the generator to align with that of the bottom-up discriminative network, leveraging the powerful discriminative representations to guide the generative model. In addition, we introduce a conditional loss that encourages the use of conditional information from the layer above, and a novel entropy loss that maximizes a variational lower bound on the conditional entropy of generator outputs. We first train each stack independently, and then train the whole model end-to-end. Unlike the original GAN that uses a single noise vector to represent all the variations, our SGAN decomposes variations into multiple levels and gradually resolves uncertainties in the top-down generative process. Based on visual inspection, Inception scores and visual Turing test, we demonstrate that SGAN is able to generate images of much higher quality  than GANs without stacking. 
\end{abstract}

\vspace{-0.15in}
\section{Introduction}
\label{intro}

Recent years have witnessed tremendous success of deep neural networks~(DNNs), especially the kind of bottom-up neural networks trained for discriminative tasks. In particular, Convolutional Neural Networks (CNNs) have achieved  impressive accuracy on the challenging ImageNet classification benchmark \cite{krizhevsky2012imagenet,simonyan2015very,szegedy2015going,he2016deep,russakovsky2015imagenet}. Interestingly, it has been shown that CNNs trained on ImageNet for classification can learn representations that are transferable to other tasks \cite{sharif2014cnn}, and even to other modalities \cite{gupta2016cross}.  However, bottom-up discriminative models are focused on learning useful representations from data, being incapable of capturing the data distribution.

Learning top-down generative models that can explain complex data distribution is a long-standing problem in machine learning research. The expressive power of deep neural networks makes them natural candidates for generative models, and several recent works have shown promising results \cite{kingma2014auto, goodfellow2014generative, oord2016pixel, li2015generative, zhao2016energy,makhzani2016adversarial,dosovitskiy2015learning}. 
While state-of-the-art DNNs can rival human performance in certain discriminative tasks, current best deep generative models still fail when there are large variations in the data distribution. 

A natural question therefore arises: can we leverage
the hierarchical representations in a 
discriminatively trained model to help the learning of top-down generative models? In this paper, we propose a generative model named {\em Stacked Generative Adversarial Networks} (SGAN). 
Our model consists of a top-down stack of GANs, each trained to generate ``plausible'' lower-level representations conditioned on higher-level representations. Similar to the image discriminator in the original GAN model which is trained to distinguish ``fake'' images from ``real'' ones, we introduce a set of \emph{representation discriminators} that are trained to distinguish ``fake'' representations from ``real'' representations. The \emph{adversarial loss} introduced by the representation discriminator forces the intermediate representations of the SGAN to lie on the manifold of the bottom-up DNN's representation space. 
In addition to the adversarial loss, we also introduce a \emph{conditional loss} that imposes each generator to use the higher-level conditional information, and a novel \emph{entropy loss} that  encourages each generator to generate diverse representations. 
By stacking several GANs in a top-down way and using the top-most GAN to receive labels and the bottom-most GAN to generate images, SGAN can be trained to model the data distribution conditioned on class labels. 
Through extensive experiments, we demonstrate that our SGAN is able to generate images of much higher quality than a vanilla GAN. In particular, our model obtains state-of-the-art Inception scores on CIFAR-10 dataset.

%
%

\section{Related Work}
\label{related}
\vspace{-0.1in}
\vpara{Deep Generative Image Models.}
There has been a large body of work on generative image modeling with deep learning. Some early efforts include Restricted Boltzmann Machines \cite{hinton2002training} and Deep Belief Networks \cite{hinton2006reducing}.
More recently, several successful paradigms of deep generative models have emerged, including the auto-regressive models \cite{larochelle2011neural,germain2015made,theis2015generative,oord2016pixel,oord2016conditional,gregor2014deep}, Variational Auto-encoders (VAEs) \cite{kingma2014auto,kingma2014semi,rezende2014stochastic,yan2016attribute2image,gregor2015draw}, and Generative Adversarial Networks (GANs) \cite{goodfellow2014generative, denton2015deep, radford2016unsupervised, reed2016generative,salimans2016improved,larsen2016autoencoding}. Our work builds upon the GAN framework, which employs a generator that transforms a noise vector into an image and a discriminator that distinguishes between real and generated images.

However, due to the vast variations in image content, it is still challenging for GANs to generate diverse images with sufficient details. 
To this end, several works have attempted to factorize a GAN into a series of GANs, decomposing the difficult task into several more tractable sub-tasks. Denton~\etal~\cite{denton2015deep} propose a LAPGAN model that factorizes the generative process  into multi-resolution GANs, with each GAN generating a higher-resolution residual conditioned on a lower-resolution image. Although both LAPGAN and SGAN consist of a sequence of GANs each working at one scale, LAPGAN focuses on generating \emph{multi-resolution images} from coarse to fine while our SGAN aims at modeling \emph{multi-level representations} from abstract to specific. 
Wang and Gupta~\cite{wang2016generative} propose a $\mathrm{S^{2}}$-GAN, using one GAN to generate  surface normals and another GAN to generate images conditioned on surface normals. Surface normals can be viewed as a specific type of image representations, capturing the underlying 3D structure of an indoor scene. On the other hand, our framework can leverage the more general and powerful multi-level representations in a pre-trained discriminative DNN. 

There are several works that use a pre-trained discriminative model to aid the training of a generator. \cite{lamb2016discriminative, dosovitskiy2016generating} add a regularization term that encourages the reconstructed image to be similar to the original image in the feature space of a discriminative network. \cite{ulyanov2016texture,johnson2016perceptual} use an additional ``style loss'' based on Gram matrices of feature activations. Different from our method, all the works above only add loss terms to regularize the generator's \emph{output}, without regularizing its \emph{internal representations}. 

\vpara{Matching Intermediate Representations Between Two DNNs.} There have been some works that attempt to ``match'' the intermediate representations between two DNNs. \cite{romero2015fitnets, gupta2016cross} use the intermediate representations of one pre-trained DNN to guide another DNN in the context of knowledge transfer. 
Our method can be considered as a special kind of knowledge transfer. However, we aim at transferring the knowledge in a bottom-up DNN  to a top-down generative model, instead of another bottom-up DNN. 
Also, some auto-encoder architectures employ layer-wise reconstruction loss~\cite{valpola2015neural,rasmus2015semi,zhao2016stacked,2016-icml-recon-dec}. 
The layer-wise loss is usually accompanied by lateral connections from the encoder to the decodery. On the other hand, SGAN is a generative model and does not require any information from the encoder once training completes. Another important difference 
is that we use adversarial loss instead of $L2$ reconstruction loss to match intermediate representations. 



\vpara{Visualizing Deep Representations.}
Our work is also related to the recent efforts in visualizing the internal representations of DNNs. One popular approach uses gradient-based optimization to find an image whose representation is close to the one we want to visualize \cite{mahendran2016visualizing}. 
Other approaches, such as \cite{dosovitskiy2016inverting}, train a top-down deconvolutional network to reconstruct the input image from a feature representation by minimizing the Euclidean reconstruction error in image space. However, there is inherent uncertainty in the reconstruction process, since the representations in higher layers of the DNN are trained to be invariant to irrelevant transformations and to ignore low-level details. With Euclidean training objective, the deconvolutional network tends to produce blurry images. 
To alleviate this problem, Dosovitskiy abd Brox~\cite{dosovitskiy2016generating} further propose a feature loss and an adversarial loss that enables much sharper reconstructions.  
However, it still does not tackle the problem of uncertainty in reconstruction. Given a high-level feature representation, the deconvolutional network deterministically generates a single image, despite the fact that there exist many images having the same representation. Also, there is no obvious way to sample images because the feature prior distribution is unknown. Concurrent to our work, Nguyen~\etal~\cite{nguyen2016plug} incorporate the feature prior with a variant of denoising auto-encoder (DAE). Their sampling relies on an iterative optimization procedure, while we are focused on efficient feed-forward sampling.

\section{Methods}
\label{methods}

In this section we introduce our model architecture. In Sec.~\ref{GAN} we briefly overview the framework of Generative Adversarial Networks. We then describe our proposal for Stacked Generative Adversarial Networks in Sec.~\ref{SGAN}. In Sect.~\ref{Condloss} and \ref{Entloss} we will focus on our two novel loss functions, conditional loss and entropy loss, respectively. 

\begin{figure*}[!htbp]
\centering
\includegraphics[width=0.9\linewidth]{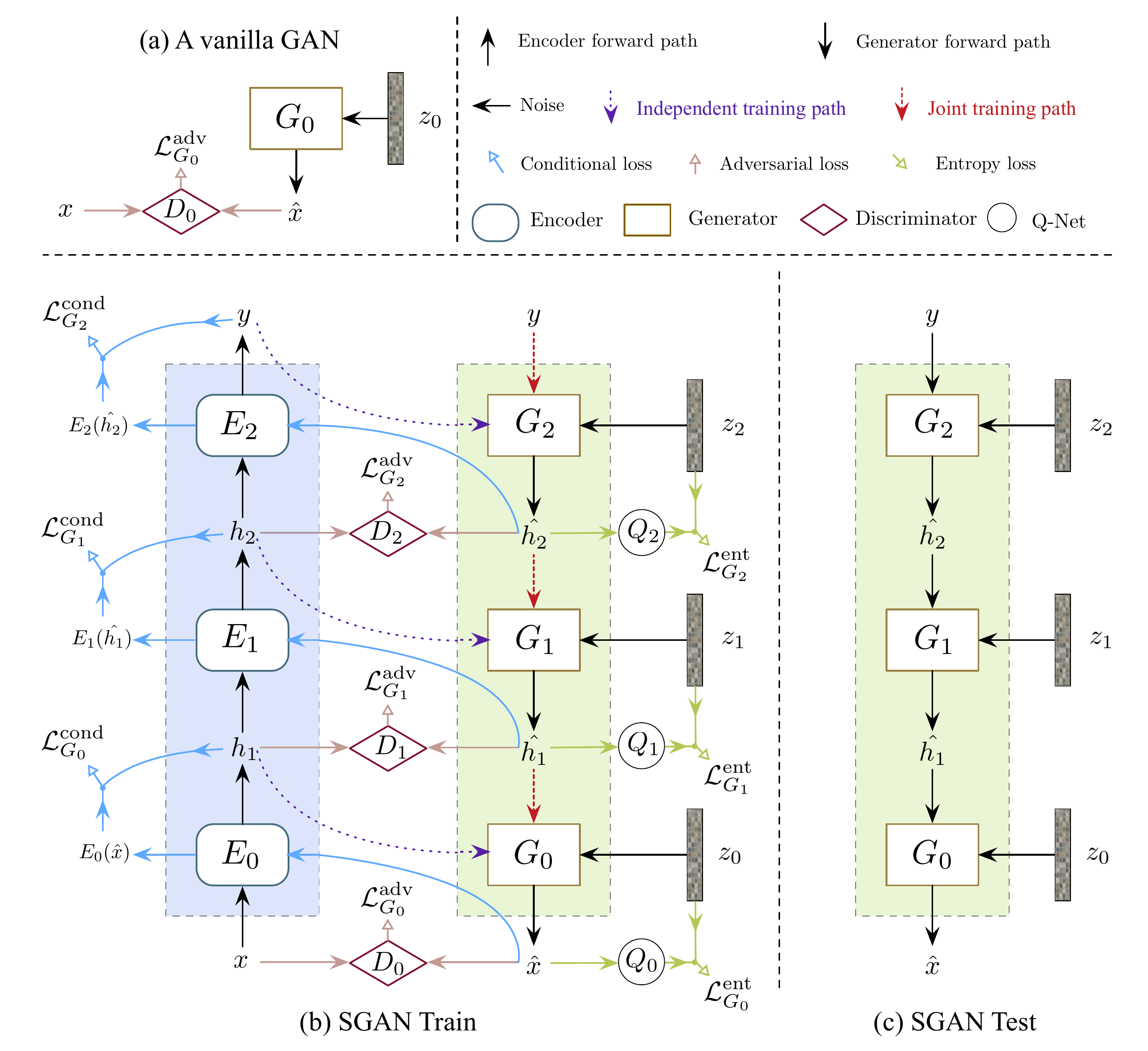}
   \caption{\textbf{An overview of SGAN.} (a) The original GAN in~\cite{goodfellow2014generative}. (b) The workflow of training SGAN, where each generator $G_i$ tries to generate plausible features that can fool the corresponding representation discriminator $D_i$. Each generator receives conditional input from encoders in the independent training stage, and from the upper generators in the joint training stage. (c) New images can be sampled from SGAN (during test time) by feeding random noise to each generator $G_i$.}
\label{fig:sgan}
\vspace{-0.3cm}
\end{figure*}

\subsection{Background: Generative Adversarial Network}
\label{GAN}
As shown in Fig.~\ref{fig:sgan}~(a), the original GAN \cite{goodfellow2014generative} is trained using a two-player min-max game: a discriminator $D$ trained to distinguish generated images from real images, and a generator $G$ trained to fool $D$. The discriminator loss $\mathcal{L}_{D}$ and the generator loss $\mathcal{L}_{G}$ are defined as follows: 
\begin{equation}
\mathcal{L}_{D} =  \mathbb{E}_{x\sim P_{data}}[-\log D(x)] +\mathbb{E}_{z\sim P_{z}}[-\log (1-D\big(G(z))\big)] 
\end{equation} 
\begin{equation}
\mathcal{L}_{G} = \mathbb{E}_{z\sim P_{z}}[ -\log (D\big(G(z))\big)] 
\end{equation}

In practice, $D$ and $G$ are usually updated alternately. The training process matches the generated image distribution $P_{G}(x)$ with the real image distribution $P_{data}(x)$ in the training set. In other words, The adversarial training forces $G$ to generate images that reside on the natural images manifold. 


\subsection{Stacked Generative Adversarial Networks}
\label{SGAN}

\vpara{Pre-trained Encoder.} We first consider a bottom-up DNN pre-trained for classification, 
which is referred to as the encoder $E$ throughout. 
We define a stack of bottom-up deterministic nonlinear mappings: $h_{i+1} = E_{i}(h_{i})$, where $i \in \{0,1,...,N-1\}$, $E_{i}$ consists of a sequence of neural layers (\emph{e.g.}, convolution, pooling), 
$N$ is the number of hierarchies (stacks), $h_{i} (i \neq 0, N)$ are intermediate representations,  $h_{N}=y$ is the classification result,  and $h_{0}=x$ is the input image.  Note that in our formulation, each $E_{i}$ can contain multiple layers and the way of grouping layers together into $E_{i}$ is determined by us. The number of stacks $N$ is therefore less than the number of layers in $E$ and is also determined by us.

\vpara{Stacked Generators.}  Provided with a pre-trained encoder $E$, our goal is to train a top-down generator $G$ that inverts $E$. Specifically, $G$ consists of a top-down \emph{stack} of generators $G_{i}$, each trained to invert a bottom-up mapping $E_{i}$. Each $G_{i}$ takes in a higher-level feature and a noise vector as inputs, and  outputs the lower-level feature $\hat{h}_{i}$. 
We first train each GAN independently and then train them jointly in an end-to-end manner, as shown in Fig.~\ref{fig:sgan}. 
Each generator receives conditional input from encoders in the independent training stage, and from the upper generators in the joint training stage.  In other words, $\hat{h}_{i} = G_{i}(h_{i+1}, z_{i})$ during independent training and $\hat{h}_{i} = G_{i}(\hat{h}_{i+1}, z_{i})$ during joint training. The loss equations shown in this section are for independent training stage but can be easily modified to joint training by replacing $h_{i+1}$ with $\hat{h}_{i+1}$. 

Intuitively, the total variations of images could be decomposed into multiple levels, with higher-level semantic variations (\emph{e.g.}, attributes, object categories, rough shapes) and lower-level variations (\emph{e.g.}, detailed contours and textures, background clutters). Our model allows using different noise variables to represent different levels of variations.


The training procedure is shown in Fig.~\ref{fig:sgan}~(b). Each generator $G_{i}$ is trained with a linear combination of three loss terms: adversarial loss, conditional loss, and entropy loss. 
\begin{equation}\mathcal{L}_{G_{i}} =\lambda_{1}\mathcal{L}_{G_{i}}^{adv} + \lambda_{2}\mathcal{L}_{G_{i}}^{cond} + \lambda_{3}\mathcal{L}_{G_{i}}^{ent},\end{equation}
where $\mathcal{L}_{G_{i}}^{adv}$, $\mathcal{L}_{G_{i}}^{cond}$, $\mathcal{L}_{G_{i}}^{ent}$ denote adversarial loss, conditional loss, and entropy loss respectively. $\lambda_{1}$, $\lambda_{2}$, $\lambda_{3}$ are the weights associated with different loss terms. In practice, we find it sufficient to set the weights such that the magnitude of different terms are of similar scales. 
In this subsection we first introduce the adversarial loss $\mathcal{L}_{G_{i}}^{adv}$. We will then introduce $\mathcal{L}_{G_{i}}^{cond}$ and $\mathcal{L}_{G_{i}}^{ent}$ in Sec.~\ref{Condloss} and  \ref{Entloss} respectively.

For each generator $G_{i}$, we introduce a \emph{representation discriminator} $D_{i}$ that distinguishes generated representations $\hat{h}_{i}$, from ``real'' representations ${h_{i}}$. Specifically, the discriminator $D_{i}$ is trained with the loss function:
\begin{multline}
\mathcal{L}_{D_{i}} = \mathbb{E}_{h_{i}\sim P_{data, E}}[-\log D_{i}\big(h_{i})]  + \\
 \mathbb{E}_{z_{i}\sim P_{z_{i}},{\ } h _{i+1}\sim P_{data, E}}[-\log \big(1-D_{i}(G_{i}(h_{i+1},z_{i}))\big)]
\end{multline}

And $G_{i}$ is trained to ``fool'' the representation discriminator $D_{i}$, with the adversarial loss defined by:
\begin{equation}\mathcal{L}_{G_{i}}^{adv} =\mathbb{E}_{h_{i+1}\sim P_{data, E},{\ } z_{i}\sim P_{z_{i}}}[-\log (D_{i}(G_{i}(h_{i+1}, z_{i})))]\end{equation}






During joint training, the adversarial loss provided by representational discriminators can also be regarded as a type of deep supervision~\cite{lee2015deeply}, providing intermediate supervision signals. In our current formulation, $E$ is a discriminative model, and $G$ is a generative model conditioned on labels. However, it is also possible to train SGAN without using label information: $E$ can be trained with an unsupervised objective and $G$ can be cast into an unconditional generative model by removing the label input from the top generator. We leave this for future exploration.



\vpara{Sampling.} To sample images, all $G_{i}$s are stacked together in a top-down manner, as shown in Fig.~\ref{fig:sgan} (c). 
Our SGAN is capable of modeling the data distribution conditioned on the class label: $p_{G}(\hat{x}|y) = p_{G}(\hat{h}_{0}|\hat{h}_{N}) \propto p_{G}(\hat{h}_{0}, \hat{h}_{1}, ..., \hat{h}_{N-1}|\hat{h}_{N}) = \prod\limits_{0\leq i \leq N-1 } p_{G_{i}}(\hat{h}_{i}| \hat{h}_{i+1})$, where each $p_{G_{i}}(\hat{h}_{i}| \hat{h}_{i+1})$ is modeled by a generator $G_{i}$. 
From an information-theoretic perspective, SGAN factorizes the total entropy of the image distribution $H(x)$ into multiple (smaller) conditional entropy terms: $H(x) = H(h_{0}, h_{1}, ..., h_{N}) = \sum_{i=0}^{N-1}H(h_{i}|h_{i+1}) + H(y)$, thereby decomposing one difficult task into multiple easier tasks.

\subsection{Conditional Loss}
\label{Condloss}
At each stack, a generator $G_{i}$ is trained to capture the distribution of lower-level representations $\hat{h}_{i}$, conditioned on higher-level representations $h_{i+1}$. However, in the above formulation,
the generator might choose to ignore $h_{i+1}$, and generate plausible $\hat{h}_{i}$ from scratch. Some previous works \cite{mirza2014conditional, gauthier2014conditional,denton2015deep} tackle this problem by feeding the conditional information to both the generator and discriminator. This approach, however, might introduce unnecessary complexity to the discriminator and increase model instability~\cite{pathak2016context,sangkloy2016scribbler}.  

Here we adopt a different approach: we regularize the generator by adding a loss term $\mathcal{L}_{G_{i}}^{cond}$ named \emph{conditional loss}. We feed the generated lower-level representations $\hat{h}_{i}=G_{i}(h_{i+1},z_{i})$ back to the encoder $E$, and compute the recovered higher-level representations. We then enforce the  recovered representations to be similar to the conditional representations. 
Formally:
\begin{equation}\mathcal{L}_{G_{i}}^{cond} = \mathbb{E}_{h_{i+1}\sim P_{data, E},{\ } z_{i}\sim P_{z_{i}}}[f(E_{i}(G_{i}(h_{i+1},z_{i})), h_{i+1})]\end{equation}
where $f$ is a distance measure. We define $f$ to be the Euclidean distance for intermediate representations 
and cross-entropy for labels. 
Our conditional loss $\mathcal{L}_{G_{i}}^{cond}$ is similar to the ``feature loss'' used by \cite{dosovitskiy2016generating} and the ``FCN loss'' in \cite{wang2016generative}. 

\subsection{Entropy Loss}
\label{Entloss}
Simply adding the conditional loss $\mathcal{L}_{G_{i}}^{cond}$ leads to another issue: the generator $G_{i}$ learns to ignore the noise $z_{i}$, and compute  $\hat{h}_{i}$ deterministically from $h_{i+1}$. 
This problem has been encountered in various applications of conditional GANs, \emph{e.g.}, synthesizing future frames conditioned on previous frames~\cite{mathieu2016deep}, generating images conditioned on label maps~\cite{pix2pix2016}, and most related to our work, synthesizing images conditioned on feature representations~\cite{dosovitskiy2016generating}. All the above works attempted to generate \emph{diverse} images/videos by feeding noise to the generator, but failed because the conditional generator simply ignores the noise. To our knowledge, there is still no principled way to deal with this issue.
It might be tempting to think that \emph{minibatch discrimination} \cite{salimans2016improved}, which encourages sample diversity in each minibatch, could solve this problem. However, even if the generator generates  $\hat{h}_{i}$ deterministically from $h_{i+1}$, the generated samples in each minibatch are still diverse since generators are conditioned on different $h_{i+1}$. Thus, there is no obvious way minibatch discrimination could penalize a collapsed conditional generator.

\vpara{Variational Conditional Entropy Maximization.} To tackle this problem, we would like to encourage the generated representation $\hat{h}_{i}$ to be sufficiently diverse when conditioned on $h_{i+1}$, \emph{i.e.}, the conditional entropy $H(\hat{h}_{i}|h_{i+1})$ should be as high as possible. Since directly maximizing $H(\hat{h}_{i}|h_{i+1})$ is  intractable, we propose to maximize instead a \emph{variational lower bound} on the conditional entropy. Specifically, we use an auxiliary distribution $Q_{i}(z_{i}|\hat{h}_{i})$ to approximate the true posterior $P_{i}(z_{i}|\hat{h}_{i})$, and augment the training objective with a loss term named \emph{entropy loss}: 
\begin{equation}
\mathcal{L}_{G_{i}}^{ent} = \mathbb{E}_{z_{i}\sim P_{z_{i}}}[\mathbb{E}_{\hat{h}_{i}\sim G_{i}(\hat{h}_{i}|z_{i})} [-\log Q_{i}(z_{i}|\hat{h}_{i})]]
\end{equation}
Below we give a proof that minimizing $\mathcal{L}_{G_{i}}^{ent}$ is equivalent to maximizing a variational lower bound for $H(\hat{h}_{i}|h_{i+1})$. 
\begin{equation}
\begin{split}
H(\hat{h}_{i}|h_{i+1}) &= H(\hat{h}_{i}, z_{i}|h_{i+1}) - H(z_{i}|\hat{h}_{i}, h_{i+1}) \\
					   &\geq  H(\hat{h}_{i}, z_{i}|h_{i+1}) - H(z_{i}|\hat{h}_{i}) \\
					   &=  H(z_{i}|h_{i+1}) + \underbrace{H(\hat{h}_{i}|z_{i},h_{i+1})}_{0}- H(z_{i}|\hat{h}_{i}) \\
					   &=  H(z_{i}|h_{i+1}) - H(z_{i}|\hat{h}_{i}) \\
					   &=  H(z_{i}) - H(z_{i}|\hat{h}_{i}) \\
        			   &=  \mathbb{E}_{\hat{h}_{i}\sim G_{i}}[\mathbb{E}_{z_{i}^{\prime}\sim P_{i}(z_{i}^{\prime}|\hat{h}_{i})} [\log P_{i}(z_{i}^{\prime}|\hat{h}_{i})]] + H(z_{i}) 	\\
       			 	   &=  \mathbb{E}_{\hat{h}_{i}\sim G_{i}}[\mathbb{E}_{z_{i}^{\prime}\sim P_{i}(z_{i}^{\prime}|\hat{h}_{i})} [\log Q_{i}(z_{i}^{\prime}|\hat{h}_{i})] \\
       			 	   &+\underbrace{KLD(P_{i} \Vert Q_{i})}_{\geq 0}] + H(z_{i}) \\
       			 	   &\geq  \mathbb{E}_{\hat{h}_{i}\sim G_{i}}[\mathbb{E}_{z_{i}^{\prime}\sim P_{i}(z_{i}^{\prime}|\hat{h}_{i})} [\log Q_{i}(z_{i}^{\prime}|\hat{h}_{i})]] + H(z_{i}) \\
       			 	   &=  \mathbb{E}_{z_{i}^{\prime}\sim P_{z_{i}^{\prime}}}[\mathbb{E}_{\hat{h}_{i}\sim G_{i}(\hat{h}_{i}|z_{i}^{\prime})} [\log Q_{i}(z_{i}^{\prime}|\hat{h}_{i})]] + H(z_{i}) \\
       			 	   &\triangleq -\mathcal{L}_{G_{i}}^{ent} + H(z_{i})
\end{split}
\end{equation}
In practice, we parameterize $Q_{i}$ with a deep network that predicts the posterior distribution of $z_{i}$ given $\hat{h}_{i}$.  $Q_{i}$ shares most of the parameters with $D_{i}$. We treat the posterior as a diagonal Gaussian with fixed standard deviations, and use the network $Q_{i}$ to only predict the posterior mean, making $\mathcal{L}_{G_{i}}^{ent}$  equivalent to the Euclidean reconstruction error. In each iteration we update both $G_{i}$ and $Q_{i}$ to minimize $\mathcal{L}_{G_{i}}^{ent}$. 

Our method is similar to the variational mutual information maximization technique proposed by Chen~\etal~\cite{chen2016infogan}.  A key difference is that \cite{chen2016infogan} uses the $Q$-network to   predict only a small set of deliberately constructed ``latent code'', while our $Q_{i}$ tries to predict \emph{all} the noise variables $z_{i}$ in each stack. The loss used in \cite{chen2016infogan} therefore maximizes the \emph{mutual information} between the output and  the latent code, while ours maximizes the \emph{entropy} of the output $\hat{h}_{i}$, conditioned on ${h}_{i+1}$. \cite{donahue2017adversarial,dumoulin2017adversarially} also train a separate network to map images back to latent space to perform unsupervised feature learning. Independent of our work, ~\cite{dai2017calibrating} proposes to regularize EBGAN~\cite{zhao2016energy} with entropy maximization in order to prevent the discriminator from degenerating to uniform prediction. Our entropy loss is motivated from generating multiple possible outputs from the same conditional input.

\section{Experiments}
\label{experiments}


In this section, we perform experiments on a variety of datasets including MNIST~\cite{lecun1998gradient}, SVHN~\cite{netzer2011reading}, and CIFAR-10~\cite{krizhevsky2009learning}.  
Code and pre-trained models are available at: \url{https://github.com/xunhuang1995/SGAN}. Readers may refer to our code repository for more details about experimental setup, hyper-parameters, \emph{etc.}	

\vpara{Encoder:} For all datasets we use a small CNN with two convolutional layers as our encoder: $\texttt{conv1-pool1-conv2-pool2-fc3-fc4}$, where $\texttt{fc3}$ is a fully connected layer 
and \texttt{fc4} outputs classification scores before softmax. 
On CIFAR-10 we apply horizontal flipping to train the encoder. No data augmentation is used on other datasets. 

\vpara{Generator:} We use generators with two stacks throughout our experiments. Note that our framework is generally applicable to the setting with multiple stacks, and we hypothesize that using more stacks would be helpful for large-scale and high-resolution datasets.  For all datasets, our top GAN $G_{1}$ generates \texttt{fc3} features from some random noise $z_{1}$, conditioned on label $y$. The bottom GAN $G_{0}$  generates images from some noise $z_{0}$, conditioned on \texttt{fc3} features generated from GAN $G_{1}$. We set the loss coefficient parameters $\lambda_1=\lambda_2=1$ and $\lambda_3=10$.\footnote{ The choice of the parameters are made so that the magnitude of each loss term is of the same scale.} 


\subsection{Datasets}
\label{datasets}

We thoroughly evaluate SGAN on three widely adopted datasets: MNIST~\cite{lecun1998gradient}, SVHN~\cite{netzer2011reading}, and CIFAR-10~\cite{krizhevsky2009learning}. The details of each dataset is described in the following. 

\vpara{MNIST:} The MNIST dataset contains $70,000$ labeled images of hand-written digits with $60,000$ in the training set and $10,000$ in the test set. Each image is sized by $28\times28$. 

\vpara{SVHN:} The SVHN dataset is composed of real-world color images of house numbers collected by Google Street View \cite{netzer2011reading}. Each image is of size $32\times32$ and the task is to classify the digit at the center of the image. 
The dataset contains $73,257$ training images and $26,032$ test images. 

\vpara{CIFAR-10:} The CIFAR-10 dataset consists of colored natural scene images sized at $32\times 32$ pixels. There are 50,000 training images and 10,000 test images in $10$ classes.  

\begin{figure}[!tbp]
	\centering
	\subfigure[SGAN samples~(conditioned on labels)]{
		\includegraphics[width=0.48\linewidth]{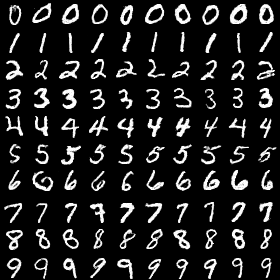}}
	\subfigure[Real images~(nearest neighbor)]{
		\includegraphics[width=0.48\linewidth]{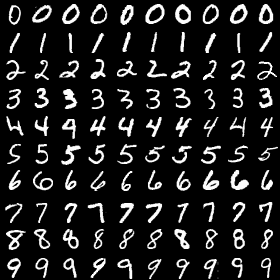}}
	\subfigure[SGAN samples~(conditioned on generated \texttt{fc3} features)]{
		\includegraphics[width=0.48\linewidth]{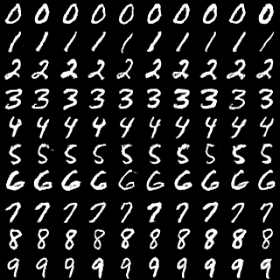}}
    \subfigure[SGAN samples~(conditioned on generated \texttt{fc3} features, trained \emph{without} entropy loss)]{
		\includegraphics[width=0.48\linewidth]{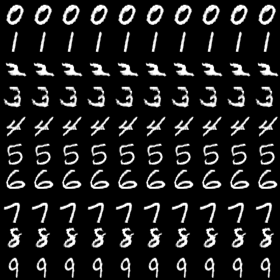}}
	\caption{{\bf MNIST results.} 
(a) Samples generated by SGAN when conditioned on class labels. (b) Corresponding nearest neighbor images in the training set. (c) Samples generated by the bottom GAN when conditioned on a fixed \texttt{fc3} feature activation, generated by the top GAN. (d) Same as (c), but the bottom GAN is trained without entropy loss.} 
	\label{fig:mnist}
	\vspace{-0.2cm}
\end{figure}

\begin{figure}[!htbp]
	\centering
	\subfigure[SGAN samples~(conditioned on labels)]{
 		\includegraphics[width=0.48\linewidth]{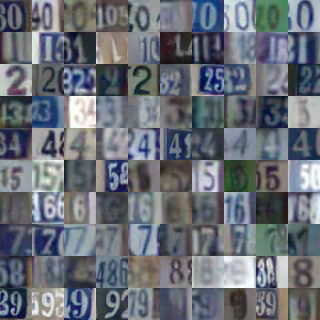}}
	\subfigure[Real images~(nearest neighbor)]{
		\includegraphics[width=0.48\linewidth]{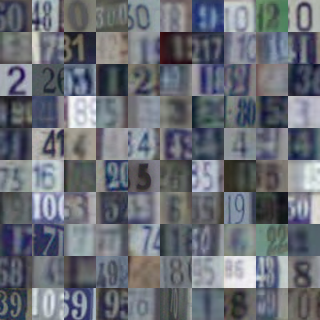}}
 	\subfigure[SGAN samples~(conditioned on generated $\texttt{fc3}$ features)]{
 		\includegraphics[width=0.48\linewidth]{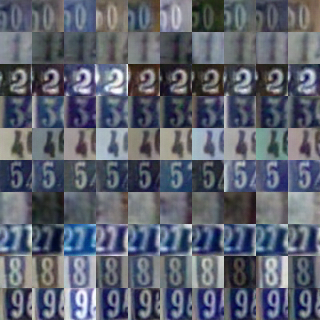}}
    \subfigure[SGAN samples~(conditioned on generated $\texttt{fc3}$ features, trained \emph{without} entropy loss)]{
		\includegraphics[width=0.48\linewidth]{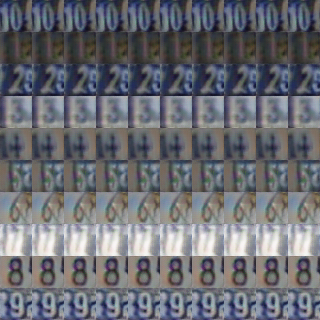}}
	\caption{{\bf SVHN results.} 
(a) Samples generated by SGAN when conditioned on class labels. (b) Corresponding nearest neighbor images in the training set. (c) Samples generated by the bottom GAN when conditioned on a fixed \texttt{fc3} feature activation, generated by the top GAN. (d) Same as (c), but the bottom GAN is trained without entropy loss.} 
 	\label{fig:svhn}
 	 \vspace{-0.2cm}
 \end{figure}
 
 \begin{figure}[!htbp]
	\centering
 	\subfigure[SGAN samples~(conditioned on labels)]{
 		\includegraphics[width=0.48\linewidth]{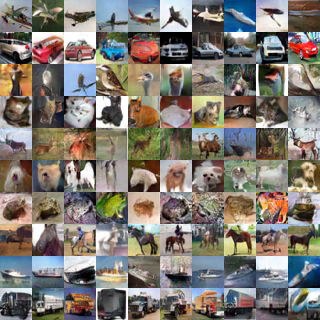}}
	\subfigure[Real images~(nearest neighbor)]{
 		\includegraphics[width=0.48\linewidth]{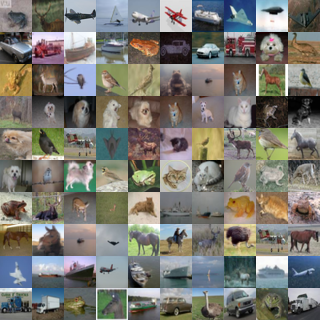}}
 	\subfigure[SGAN samples~(conditioned on generated $\texttt{fc3}$ features)]{
 		\includegraphics[width=0.48\linewidth]{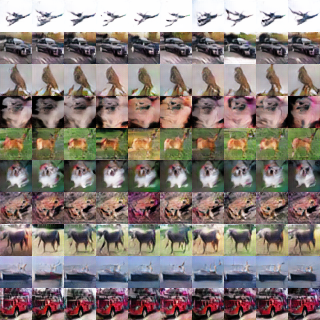}}
    \subfigure [SGAN samples~(conditioned on generated $\texttt{fc3}$ features, trained \emph{without} entropy loss)]{
		\includegraphics[width=0.48\linewidth]{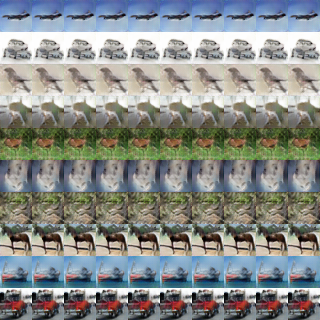}}
	\caption{{\bf MNIST results.} 
(a) Samples generated by SGAN when conditioned on class labels. (b) Corresponding nearest neighbor images in the training set. (c) Samples generated by the bottom GAN when conditioned on a fixed \texttt{fc3} feature activation, generated by the top GAN. (d) Same as (c), but the bottom GAN is trained without entropy loss.} 
 	\label{fig:cifar}
 	 \vspace{-0.2cm}
 \end{figure}
 
\subsection{Samples}
\label{samples}

In Fig.~\ref{fig:mnist}~(a), we show MNIST samples generated by SGAN. Each row corresponds to samples conditioned on a given digit class label. SGAN is able to generate diverse images with different characteristics. The samples are visually indistinguishable from real MNIST images shown in Fig.~\ref{fig:mnist}~(b), but still have differences compared with corresponding nearest neighbor training images. 

We further examine the effect of entropy loss. In Fig.~\ref{fig:mnist} (c) we show the samples generated by bottom GAN when conditioned on a fixed \texttt{fc3} feature generated by the top GAN. 
The samples (per row) have sufficient low-level variations, which reassures that bottom GAN learns to generate images without ignoring the noise $z_0$. 
In contrast, in Fig.~\ref{fig:mnist}~(d), we show samples generated without using entropy loss for bottom generator, where we observe that the bottom GAN ignores the noise and instead deterministically generates images from \texttt{fc3} features. 

An advantage of SGAN compared with a vanilla GAN is its interpretability: it decomposes the total variations of an image into different levels. For example, in MNIST it decomposes the variations into $y$ that represents the high-level digit label, $z_{1}$ that captures the mid-level coarse pose of the digit and $z_{0}$ that represents the low-level spatial details. 

The samples generated on SVHN and CIFAR-10 datasets can be seen in Fig.~\ref{fig:svhn} and Fig.~\ref{fig:cifar}, respectively. Provided with the same $\texttt{fc3}$ feature, we see in each row of panel (c) that SGAN is able to generate samples with similar coarse outline but different lighting conditions and background clutters. Also, the nearest neighbor images in the training set indicate that SGAN is not simply memorizing training data, but can truly generate novel images. 

\subsection{Comparison with the state of the art}
Here, we compare SGAN with other state-of-the-art generative models on CIFAR-10 dataset. The visual quality of generated images is measured by the widely used metric, Inception score~\cite{salimans2016improved}. Following~\cite{salimans2016improved}, we sample $50, 000$ images from our model and use the code provided by~\cite{salimans2016improved} to compute the score. As shown in Tab.~\ref{tab:inception}, SGAN obtains a score of $8.59\pm 0.12$, outperforming AC-GAN~\cite{odena2017conditional} ($8.25\pm 0.07$) and Improved GAN~\cite{salimans2016improved} ($8.09\pm 0.07$). Also, note that the $5$ techniques introduced in~\cite{salimans2016improved} are not used in our implementations. Incorporating these techniques might further boost the performance of our model.

\begin{table}[!tbp]
\centering
\begin{threeparttable}
\renewcommand{\arraystretch}{1.2}
\begin{tabular}{lllll}
\hline
Method          		& Score \\ \hline
Infusion training~\cite{bordes2017learning} & $4.62\pm 0.06$ \\ 
ALI~\cite{dumoulin2017adversarially} (as reported in~\cite{david2017improving}) 				& $5.34\pm 0.05$ \\ 
GMAN~\cite{durugkar2017gman} (best variant)	& $6.00\pm 0.19$ \\ 
EGAN-Ent-VI~\cite{dai2017calibrating} 					& $7.07\pm 0.10$ \\
LR-GAN~\cite{yang2017lrgan}	& $7.17\pm 0.07$ \\ 
Denoising feature matching~\cite{david2017improving} 					& $7.72\pm 0.13$ \\ \hline
$\text{DCGAN}^{\dag}$ (with labels, as reported in ~\cite{wang2017learning}) 		& $6.58$ \\ 
$\text{SteinGAN}^{\dag}$~\cite{wang2017learning} 		& $6.35$ \\ 
$\text{Improved GAN}^{\dag}$~\cite{salimans2016improved} (best variant)	& $8.09\pm 0.07$ \\ 
$\text{AC-GAN}^{\dag}$~\cite{odena2017conditional}			& $8.25\pm 0.07$ \\ \hline
DCGAN~$(\mathcal{L}^{adv})$	& $6.16\pm 0.07$ \\ 
DCGAN~$(\mathcal{L}^{adv}+\mathcal{L}^{ent})$	& $5.40\pm 0.16$ \\ 
DCGAN~$(\mathcal{L}^{adv}+\mathcal{L}^{cond})^{\dag}$	& $5.40\pm 0.08$ \\ 
DCGAN~$(\mathcal{L}^{adv}+L^{cond}+\mathcal{L}^{ent})^{\dag}$			& $7.16\pm 0.10$ \\ 
$\textbf{SGAN-no-joint}^{\dag}$	& $\textbf{8.37} \pm 0.08$ \\
$\textbf{SGAN}^{\dag}$		& $\textbf{8.59} \pm 0.12$ \\ \hline
Real data 				& $11.24\pm 0.12$ \\ 
\end{tabular}
\begin{tablenotes}
\item [\dag] Trained with labels.
\end{tablenotes}
\caption{\textbf{Inception Score on CIFAR-10.} SGAN and SGAN-no-joint outperform previous state-of-the-art approaches.}\label{tab:inception}
\end{threeparttable}
\vspace{-0.2cm}
\end{table}

\subsection{Visual Turing test}
To further verify the effectiveness of SGAN, we conduct human visual Turing test in which we ask AMT workers to distinguish between real images and images generated by our networks. We exactly follow the interface used in Improved GAN~\cite{salimans2016improved}, in which the workers are given $9$ images at each time and can receive feedback about whether their answers are correct. With $9,000$ votes for each evaluated model, our AMT workers got $24.4\%$ error rate for samples from SGAN and $15.6\%$ for samples from DCGAN~$(\mathcal{L}^{adv}+L^{cond}+\mathcal{L}^{ent})$. This further confirms that our stacked design can significantly improve the image quality over GAN without stacking.


\subsection{More ablation studies}
In Sec.~\ref{samples} we have examined the effect of entropy loss. In order to further understand the effect of different model components, we conduct extensive ablation studies by evaluating several baseline methods on CIFAR-10 dataset. If not mentioned otherwise, all models below use the same training hyper-parameters as the full SGAN model.
\begin{enumerate}[(a)]
\item SGAN: The full model, as described in Sec.~\ref{methods}.
\item SGAN-no-joint: Same architecture as (a), but each GAN is trained \emph{independently}, and there is no final joint training stage. 
\item DCGAN~($\mathcal{L}^{adv}+\mathcal{L}^{cond}+\mathcal{L}^{ent}$): This is a \emph{single} GAN model with the same architecture as the bottom GAN in SGAN, except that the generator is conditioned on labels instead of \texttt{fc3} features. Note that other techniques proposed in this paper, including conditional loss $\mathcal{L}^{cond}$ and entropy loss $\mathcal{L}^{ent}$, are still employed. We also tried to use the full generator $G$ in SGAN as the baseline, instead of only the bottom generator $G_{0}$. However, we failed to make it converge, possibly because $G$ is too deep to be trained without intermediate supervision from representation discriminators.
\item DCGAN~($\mathcal{L}^{adv}+\mathcal{L}^{cond}$): Same architecture as (c), but trained without entropy loss $\mathcal{L}^{ent}$.
\item DCGAN~($\mathcal{L}^{adv}+\mathcal{L}^{ent}$): Same architecture as (c), but trained without conditional loss $\mathcal{L}^{cond}$. This model therefore does not use label information.
\item DCGAN~($\mathcal{L}^{adv}$): Same architecture as (c), but trained with neither conditional loss $\mathcal{L}^{cond}$ nor entropy loss $\mathcal{L}^{ent}$. This model also does not use label information. It can be viewed as a plain unconditional DCGAN model~\cite{radford2016unsupervised} and serves as the ultimate baseline.
\end{enumerate}

 \begin{figure}[!tbp]
	\centering
 		\subfigure[SGAN]{
 		\includegraphics[width=0.48\linewidth]{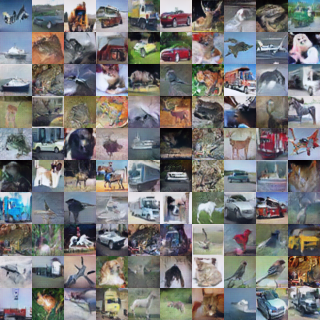}}        
 		\subfigure[SGAN-no-joint]{
 		\includegraphics[width=0.48\linewidth]{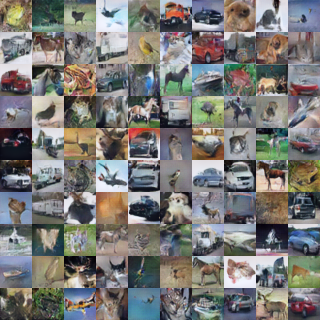}}        
    	\subfigure [\hspace{-0.06cm}DCGAN\hspace{0.03cm}($\mathcal{L}^{adv}+\mathcal{L}^{cond}+\mathcal{L}^{ent})$]{
		\includegraphics[width=0.48\linewidth]
        {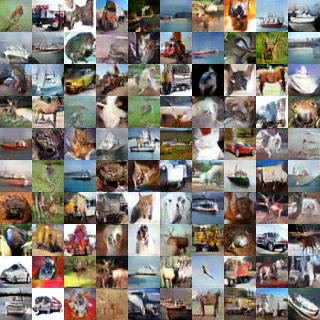}}        
    	\subfigure [DCGAN~($\mathcal{L}^{adv}+\mathcal{L}^{cond}$)]{
		\includegraphics[width=0.48\linewidth]
        {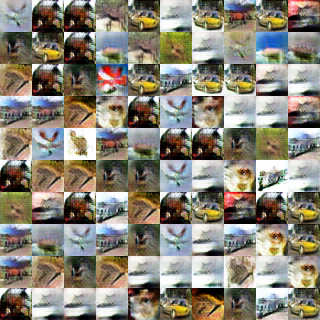}}        
    	\subfigure [DCGAN~($\mathcal{L}^{adv}+\mathcal{L}^{ent}$)]{
		\includegraphics[width=0.48\linewidth]
        {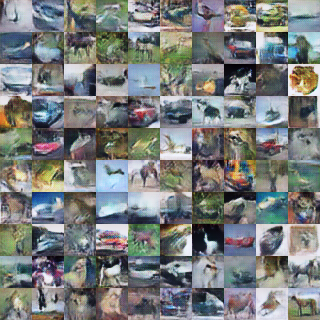}}        
        \subfigure [DCGAN~($\mathcal{L}^{adv}$)]{
		\includegraphics[width=0.48\linewidth]
        {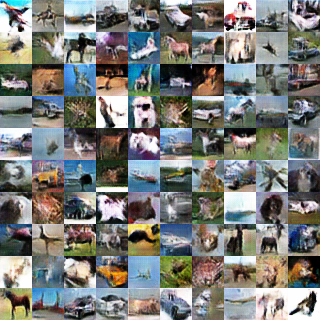}}
 	\caption{{\bf Ablation studies on CIFAR-10.} Samples from (a) full SGAN (b) SGAN without joint training. (c) DCGAN trained with $\mathcal{L}^{adv}+\mathcal{L}^{cond}+\mathcal{L}^{ent}$ (d) DCGAN trained with $\mathcal{L}^{adv}+\mathcal{L}^{cond}$ (e) DCGAN trained with $\mathcal{L}^{adv}+\mathcal{L}^{ent}$ (f) DCGAN trained with $\mathcal{L}^{adv}$. } 
 	\label{fig:ablation}
 	 \vspace{-0.2cm}
 \end{figure}

We compare the generated samples~(Fig.~\ref{fig:ablation}) and Inception scores~(Tab.~\ref{tab:inception}) of the baseline methods. Below we summarize some of our results:
\begin{enumerate}[1)]
\item SGAN obtains slightly higher Inception score than SGAN-no-joint. 
Yet SGAN-no-joint also generates very high quality samples and outperforms all previous methods in terms of Inception scores. 
\item SGAN, either with or without joint training, achieves significantly higher Inception score and better sample quality than the baseline DCGANs. This demonstrates the effectiveness of the proposed stacked approach.
\item As shown in Fig.~\ref{fig:ablation}~(d), DCGAN~($\mathcal{L}^{adv}+\mathcal{L}^{cond}$) collapses to generating a single image per category, while adding the entropy loss enables it to generate diverse images (Fig.~\ref{fig:ablation}~(c)). This further demonstrates that entropy loss is effective at improving output diversity.
\item The single DCGAN ($\mathcal{L}^{adv}+\mathcal{L}^{cond}+\mathcal{L}^{ent}$) model obtains higher Inception score than the conditional DCGAN reported in~\cite{wang2017learning}. This suggests that $\mathcal{L}^{cond}+\mathcal{L}^{ent}$ might offer some advantages compared to a plain conditional DCGAN, even without stacking.
\item In general, Inception score~\cite{salimans2016improved} correlates well with visual quality of images. However, it seems to be insensitive to diversity issues							. For example, it gives the same score to Fig.~\ref{fig:ablation}~(d) and (e) while (d) has clearly collapsed. This is consistent with results in~\cite{odena2017conditional,wang2017learning}.
\end{enumerate}

\section{Discussion and Future Work}
\label{discussion}

This paper introduces a top-down generative framework named SGAN, which effectively leverages the representational information from a pre-trained discriminative network. Our approach decomposes the hard problem of estimating image distribution into multiple relatively easier tasks -- each generating plausible representations conditioned on higher-level representations. The key idea is to use representation discriminators at different training hierarchies to provide intermediate supervision. We also propose a novel entropy loss to tackle the problem that conditional GANs tend to ignore the noise. Our entropy loss could be employed in other applications of conditional GANs, \emph{e.g.}, synthesizing \emph{different} future frames given the same past frames~\cite{mathieu2016deep}, or generating a \emph{diverse} set of images conditioned on the same label map~\cite{pix2pix2016}. We believe this is an interesting research direction in the future.



\subsubsection*{Acknowledgments}

We would like to thank Danlu Chen for the help with Fig.~\ref{fig:sgan}. Also, we want to thank Danlu Chen, Shuai Tang, Saining Xie, Zhuowen Tu, Felix Wu and Kilian  Weinberger for helpful discussions.  Yixuan Li is supported by US Army Research Office W911NF-14-1-0477. Serge Belongie is supported in part by a Google Focused Research Award.

{\small
\bibliographystyle{ieee}
\bibliography{egbib}
}

\end{document}